# Lévy walks derived from a Bayesian decision-making model in non-stationary environments

**Shuji Shinohara[1*], Nobuhito Manome[12], Yoshihiro Nakajima[3], Yukio Pegio Gunji[4], Toru Moriyama[5], Hiroshi Okamoto[1], Shunji Mitsuyoshi[1], Ung-il Chung[1]**

1. Department of Bioengineering, Graduate School of Engineering, The University of Tokyo, Tokyo, 113-8656, Japan
2. Department of Research and Development, SoftBank Robotics Group Corp., Tokyo, 105-0021, Japan
3. Graduate School of Economics, Osaka City University, Osaka, 558-8585, Japan
4. Department of Intermedia Art and Science, School of Fundamental Science and Technology, Waseda University, Tokyo, 169-8555, Japan
5. Faculty of Textile Science, Shinshu University, Ueda, 386-8567, Japan

* Corresponding author E-mail: shinohara@bioeng.t.u-tokyo.ac.jp

**Abstract**

Lévy walks are found in the migratory behaviour patterns of various organisms, and the reason for this phenomenon has been much discussed. We use simulations to demonstrate that learning causes the changes in confidence level during decision-making in non-stationary environments, and results in Lévy-walk-like patterns.

One inference algorithm involving confidence is Bayesian inference. We propose an algorithm that introduces the effects of learning and forgetting into Bayesian inference, and simulate an imitation game in which two decision-making agents incorporating the algorithm estimate each other's internal models from their opponent's observational data.

For forgetting without learning, agent confidence levels remained low due to a lack of information on the counterpart and Brownian walks occurred for a wide range of forgetting rates. Conversely, when learning was introduced, high confidence levels occasionally occurred even at high forgetting rates, and Brownian walks universally became Lévy walks through a mixture of high- and low-confidence states.

## I. INTRODUCTION

Lévy walks are found in the migratory behaviour of organisms ranging from bacteria and T-cells to humans (1-7) (Figure 1A). In contrast, the movement of fine particles suspended in a liquid or gas generally follows Brownian walks (Figure 1B). Although both Lévy walks and Brownian walks are types of random walks, the frequency of occurrence of step length $l$ is characterized by the power law distribution $P(l) \sim l^{-\eta}$ , where $1<\eta\leq 3$, in a Lévy walk, while that occurring in a Brownian walk follows an exponential distribution $P(l) \sim e^{-\lambda l}$. In other words, the former is sometimes accompanied by linear movements over long distances compared to the latter, and the reason for this pattern in the migration of organisms has been the subject of much discussion (8).

One hypothesis that explains this phenomenon is the Lévy walk foraging hypothesis (9, 10). It is hypothesised that if the prey is sparse and randomly scattered and the predator has no information (memory) about the prey, a Lévy walk will be the optimal foraging behaviour (11) and will be evolutionarily advantageous.

However, humans, for example, have advanced learning and inference abilities, and can obtain useful information by performing systematic searches even in dynamic and uncertain environments (12). Rhee et al. state that human walks are not random walks, but the patterns of human walks and Lévy walks contain some statistical similarity (13). Reynolds also states that Lévy walk movement patterns could, in principle, result from non-random foraging behaviours when intelligent individuals make heavy use of cognitive properties such as mental maps, memory, and complex decision-making (7). It is also known that individual human movements are strongly influenced by the collective mobility behaviour of other people (14). Thus, at least in organisms with advanced cognitive abilities, Lévy walks may emerge as a result of decision-making through interactions with the environment and other individuals, rather than as a result of a random search (12, 15, 16).

When we perform exploratory behaviour based on decision-making, metaphorically speaking, we explore the periphery in detail if we think ‘it might be around here’ (Brownian walk). On the other hand, if we are convinced that ‘it's not around here’, we give up the search and move to a different, more distant location. In other words, a Lévy walk may result from a combination of a stray search and a conviction-based linear walk (Figure 1C). Therefore, we hypothesise that changes in confidence in one's chosen behaviour can cause Lévy-walk-like movement patterns.

One probabilistic estimation algorithm that deals with confidence is Bayesian inference. The Bayesian inference process is similar to a scientific verification procedure. In Bayesian inference, we first prepare several hypotheses (generative models) for the estimation target. We then update the confidence in each hypothesis by observing the evidence (data generated from the estimated target) and quantitatively assessing the fit between each hypothesis and the observed data. Finally, we narrow down the best hypothesis to one based on confidence.

Consider an experiment trying to estimate the probability of getting ‘1’ when rolling a die. Bayesian inference estimates that if you roll the die six times and get a ‘1’ only once, the probability of getting ‘1’ is 1/6. Similarly, Bayesian inference estimates that if you roll the die 60000 times and get ‘1’ 10000 times, the probability of getting ‘1’ is 1/6. However, the confidence in the hypothesis that ‘the probability of getting “1” is 1/6’ is higher in the latter case, as it has more trials. Thus, in Bayesian inference, the amount of information obtained from an estimation target is linked to the confidence in the estimation, and confidence is a measure of the accuracy of the estimation. In general, to accurately estimate a target, it is better to have as much observational data (information) as possible. However, this is true only for a stationary environment. When the probability of a die changes dynamically, or a die is replaced by another die in the middle of the process, it is necessary to truncate the observed data (information) in the distant past for a more accurate estimation. Therefore, as long as we use Bayesian inference in a non-stationary environment, we are forced to make decisions with limited information. Metaphorically, because we cannot be confident that ‘it's not around here’ in non-stationary environments, we will continue to explore the surrounding area by a Brownian walk. Conversely, for a Lévy walk to appear in a non-stationary environment, behaviour based on high conviction must occasionally appear (Figure 1C). Thus, the question we should ask in this study is: ‘What kind of mechanism makes it possible to be convinced of something despite a lack of information about the target of the search?’

Note the following two points regarding Bayesian inference. First, hypotheses that are candidates for the correct answer must be prepared prior to the inference. Second, the content of the hypothesis should not be changed during inference. These are natural requests from the perspective of scientific verification. However, even if we apply known hypotheses to an object seen for the first time, there may not be a correct answer among them. In such cases, it is necessary to make new hypotheses. Peirce proposed the idea of abduction (hypothesis formation) as a third form of reasoning following deduction and induction (17, 18). Arrechi et al. (19) proposed the idea of inverse Bayesian inference, in which a hypothesis given beforehand in Bayesian inference is formed from observational data.

Gunji et al. (20, 21) proposed a new mathematical model called Bayesian and inverse Bayesian (BIB) inference, which performs Bayesian and inverse Bayesian inference at the same time. Furthermore, Gunji et al. (22) proposed a swarm model that incorporates BIB inference into the self-propelled particle model and showed that Lévy walks emerge universally in swarming behaviours through simulations using the model. Horry et al. (23) showed that human decision-making can be modelled effectively by BIB inference. Shinohara et al. (24, 25) proposed an extended Bayesian inference model that incorporates the effects of forgetting and learning (inverse Bayesian inference) by introducing causal inference into Bayesian inference. Extended Bayesian inference makes it possible to adapt to dynamic environments by forgetting, and it becomes possible to deal with unknown objects by learning.

In response to the previously posed question, we show that certainty despite limited information is possible by inverse Bayesian inference (learning). Specifically, we first build a decision-making agent with the extended Bayesian inference, which introduces the effects of learning and forgetting into the agent's decision-making process. Next, we use two decision-making agents to simulate an imitation game in which each agent estimates the other's internal generative models from the other's output data.

The results show that a Brownian walk-like pattern emerges among agents who use the forgetting effect only, due to persistently low confidence and fluctuation in the estimation results. On the other hand, we show that among agents who use learning in addition to forgetting, a high degree of confidence is often achieved despite a lack of information, resulting in the emergence of Lévy-walk-like patterns.

## II. RESULTS

### Simulation results

Figure 2 shows a comparison of the simulation results for the no-learning case (γ = 0.0) and learning case (γ = 0.1) with forgetting rate β=0.3. Figures 2A, 2B, and 2C show the normal distribution mean value estimated by each agent, the hypothesis $h^t_{\max}$ that Agent1 believes most, and Agent1's confidence in each hypothesis, respectively. In the case without learning (γ = 0.0), the confidence never stays near 1 throughout the simulation period, and $h^t_{\max}$ changes frequently. On the other hand, we can see that the overall confidence level is higher with the introduction of learning, and that there are periods when the confidence level is almost 1 and $h^t_{\max}$ is stable.

For clarity, the time evolution of $h^t_{\max}$, as shown in Figure 2B, was replaced with a two-dimensional movement pattern $\left(x^t, y^t\right)$ as follows:

$$\begin{aligned} x^t &= x^{t-1} + C\cos\left(\theta^t\right) \\ y^t &= y^{t-1} + C\sin\left(\theta^t\right) \\ \theta^t &= \begin{cases} 2\pi \times rnd^t & if\ h^t_{\max} \neq h^{t-1}_{\max} \\ \theta^{t-1} & otherwise \end{cases} \end{aligned} \qquad [1]$$

where $rnd^t$ is a uniform random number generated in the range [0.0, 1.0] at time *t*. *C* is a parameter that represents the distance to be travelled in one step and is set to 1. $\theta^t$ is a variable that represents the direction of travel and changes direction randomly when $h^t_{max}$ changes. If $h^t_{max}$ does not change, $\theta^t$ will not change, and so the pattern will continue in the same direction. The results are shown in Figures 3A (γ=0.0) and 3C (γ=0.1). Figures 3B and 3D show the cumulative distribution functions (CDF) of $l$, where $l$ is the linear distance from changing direction at one point to changing direction at the next point. These figures also show the results of fitting a truncated power law distribution model (green) and an exponential distribution model (red) to the simulation data. Of the observed data, we use only data within the range of $\hat{l}_{min} \leq l \leq \hat{l}_{max}$ to improve the fitting as much as possible. See the SI Appendix for details on how to determine the fitting range $\left[\hat{l}_{min}, \hat{l}_{max}\right]$ and the fitting method. The CDF graph is such that the CDF value when $l = \hat{l}_{min}$ is set to 1. This is also true in the following CDF graphs. To analyse the frequency distribution of $l$, we performed 1000 simulation trials with different random seeds and used the aggregated data.

It can be seen from Figure 3B that the frequency distribution of $l$ is characterized by an exponential distribution. In other words, the movement patterns shown in Figure 3A are Brownian walks. The exponents for β=0.3, 0.5, and 0.7 were λ=0.24, 0.42, and 0.60, respectively. The larger β is, the more observational data (information) is truncated, and the larger the value of λ.

Figure 3D shows that the CDF of $l$ is characterized by a truncated power law distribution. The exponents for β=0.3, 0.5, and 0.7 are η=1.73, 2.00, and 2.44, respectively, and satisfy $1 < \eta \leq 3$ in all cases. That is, the movement patterns shown in Figure 3C are all Lévy walks. However, as the forgetting rate increases, the exponents increase because of the information deficit. As can be seen by comparing Figures 3B and 3D, introducing the effect of learning (inverse Bayesian inference) (γ>0.0) turns a Brownian walk into a Lévy walk.

Originally, the extended Bayesian inference model has a single parameter, as shown in the SI Appendix and references (24, 25). That is, we treated only the cases of β=γ and represent these as the same parameter α. Figure 4 shows the CDF of $l$ in the cases of α=0.3, 0.5, and 0.7. These results represent all Lévy walks with the exponents close to 2.0. The movement pattern for α=0.0, i.e., for Bayesian inference (β=γ=0.0), is shown in Figure 5A. Figures 5B and 5C show that the movement pattern turns into a Lévy walk by setting β to 0.005.

**Comparison between with and without learning**

We analysed how confidence changes with changes in forgetting rates. Figure 6A shows the relationship between the time averages of the confidence values in $h^t_{max}$ and the forgetting rate. With or without the learning effect, the mean value of confidence decreases as the forgetting rate increases in both cases. In other words, as non-stationarity increases and the amount of information decreases, the confidence decreases. However, a comparison at the same forgetting rate shows that the presence of learning increases the confidence level.

Figure 6B shows the relationship between confidence and linear step length. The confidence shown here is the average confidence during a linear move of a certain length *l*. In the absence of learning ($\gamma$=0.0), the maximum step length was 60. The maximum confidence level was approximately 0.40, although it tended to increase as the step length increased. In other words, the confidence level is not very high, and we do not see a long movement in terms of step length. Conversely, in the case with learning ($\gamma$=0.1), the confidence increases as the step length increases, and the confidence exceeds 0.99 when the step length is approximately 300. Thus, agents travel long distances with strong confidence. To calculate the average confidence in Figures 6A and 5B, we performed 1000 trials of simulations with different random seeds and used the aggregated data.

## III. DISCUSSION

In this study, we hypothesise that in organisms such as humans that have a high degree of reasoning and learning ability, changes in confidence in one's chosen behaviour can cause Lévy-walk-like patterns of behaviour. To test this hypothesis, we consider the question ‘What mechanism allows us to be convinced of something even though there is insufficient information for the search target?’ To deal with this problem, we first proposed an extended Bayesian inference that introduces forgetting and learning (inverse Bayesian inference) functions into Bayesian inference. We then incorporated this extended Bayesian inference into two decision-making agents and simulated an imitation game in which they read each other's internal states.

We choose an action from among various options in our daily decision-making, and this choice is accompanied by a degree of certainty. Even if the same action A is selected as a result in different cases, the degree of confidence at the time of selection varies, for example, thinking ‘only A is possible’ or ‘it may be A’. When we lack information and experience, we cannot be sure of our choices. Conversely, when one makes a choice based on complete information, one has confidence in the choice. In other words, the amount of information obtained is proportional to the degree of confidence. Bayesian inference deals with such confidence. However, an action chosen with certainty can actually be wrong. This type of confidence stems from our own active assumptions, rather than objective and complete information. In this study, we model such active assumptions as inverse Bayesian inference.

In stationary environments, Bayesian inference, which utilizes all information from the distant past to the present, is the best method of inference. However, for proper estimation in non-stationary environments, it is necessary to discard information about the distant past. In this case, the agent is not sure of its estimation results because of the lack of information. As a result, it frequently changes the estimates, and a Brownian walk appears.

As the lack of information becomes more pronounced, that is, the rate of forgetting increases, the exponent of the exponential distribution that the frequency distribution of the step length follows increases. By introducing inverse Bayesian inference, the frequency distributions of the step length change from exponential distributions to truncated power law distributions. That is, the Brownian walk turns into a Lévy walk regardless of the degree of information deficiency. Interestingly, in the cases where the forgetting rate and the learning rate coincide, over a wide range of parameter areas, the exponents of the truncated power law distributions are close to 2.0, which is the value that achieves optimal foraging behaviour (11).

In inverse Bayesian inference, confidence is increased by modifying the content of the hypothesis to fit the observed data. As shown in Figures 2C and Figure 6A, inverse Bayesian inference (learning) has the effect of increasing the confidence level, and even in situations where there is insufficient information, the confidence level sometimes approaches 1. In this case, as shown in Figure 6B, a longer step length tends to correspond to higher confidence. The mixture of high- and low-confidence states leads to the Lévy walk shown in Figure 3C.

Although there is controversy over whether the Lévy walk foraging hypothesis is correct or whether Lévy walks emerge as a result of interactions with the environment and other individuals (26), the results of this study do not reject the Lévy walk foraging hypothesis. Rather, they suggest that changes in confidence in decision-making can also cause Lévy-walk-like movement patterns.

In the Exploration and Preferential Return (EPR) model, both exploration as a random walk process and the human propensity to revisit places that we have visited before (preferential return) are incorporated. One of these two competing mechanisms is selected probabilistically at each time step in EPR (14). Exploration and exploitation are two essential components of any optimization algorithm (27). Finding an appropriate balance between exploration and exploitation is the most challenging task in the development of any meta-heuristic algorithm because of the randomness of the optimization process (28). In our simulations, a Lévy walk was produced by a mixture of two competing behaviours: a long-distance linear movement based on confidence and a short-distance movement performed while lost. Interestingly, these conflicting patterns of behaviour arise from a single system of extended Bayesian inference without being prepared separately.

Namboodiri et al. (12) showed through experiments with humans that the discount rate (forgetting rate) plays an important role in search behaviour. As shown in Figures 3 and 5, migration patterns are characterized by forgetting rates in our model. When the forgetting rate is zero, as shown in Figure 5A, there is no information loss, and the movement pattern is a ballistic pattern based on beliefs grounded in complete information. Conversely, when the forgetting rate is high, the estimation results (the hypothesis with maximum confidence) fluctuate because the information is truncated and the confidence level decreases. As a result, a Brownian walk appears, as shown in Figure 3A. Lévy walks, which combine both properties, appear in the parameter region between the ballistic pattern and the Brownian walk (Figure 5B).

Abe (29) constructed a model in which the Lévy walk appears by coupling two tent maps. In this model, the coupling strength between the two tent maps is represented by the parameter ε. For large values of ε, the output values $x^t$ and $y^t$ of the two coupling tent maps become identical, and the movement pattern in the 2-dimensional coordinates becomes a straight line. Conversely, when ε is small, these move independently, and the movement pattern is a Brownian walk. A Lévy walk appears in a narrow parameter range between the two extremes.

In our model, when the forgetting rate is small, the estimates between the two agents tend to agree and have higher confidence as they acquire more information about each other. In contrast, when the forgetting rate is high, the confidence is lowered due to the lack of information about the other party, and inconsistency occurs between the two estimates. Thus, the forgetting rate can be interpreted as a parameter representing the coupling strength between two agents. A simple comparison is not possible because of the difference between tent maps and Bayesian inference. However, we may be able to discuss self-organizing criticality (SOC) in our model, as Abe did, by mapping β to ε. In the case of SOC, Lévy walks are found in a very narrow parameter range. Hence, in order to explain the universality of Lévy walks found in nature, it is necessary to explain why such parameter regions are selected. Abe attributed this phenomenon to the large dynamic range (29). In contrast, in our model, Lévy walks are universally found in a wide range of forgetting rate regions by introducing inverse Bayesian inference (learning). That is, in our model, the argument for SOC is only valid in the special case where the learning rate is $\gamma$=0.0. Using a swarm model incorporating BIB inference, Gunji et al. showed that critical phenomena universally emerge without SOC (22). In this model, critical behaviour is achieved not only at the edge of chaos, but anywhere in the parameter space. Detailed discussions on SOC using our model will be added in the future.

## IV. Conclusions

In this study, we demonstrated by simulation that changes in confidence during decision-making cause Lévy-walk-like behavioural patterns. We proposed an algorithm that introduced the effects of learning and forgetting into Bayesian inference, and simulated an imitation game between decision-making agents that incorporated the algorithm.

In the case of only forgetting without learning, confidence level in one's chosen behaviour remained low due to a lack of the other party information, and frequently changed the estimation results. As a result, Brownian-walk-like behavioural patterns were observed in a wide range of forgetting-rate areas.

In the case of neither forgetting nor learning, that is, in the case of Bayesian inference, the other party information was not truncated, so that the confidence level remained high based on the complete information and the behaviour pattern became ballistic. Lévy-walk-like behavioural patterns appeared only in forgetting rate regions near zero.

Conversely, when learning was introduced, Lévy-walk-like behavioural patterns emerged universally in a wide range of parameter areas by the mixture of high- and low-confidence states. The results suggest that changes in confidence in decision-making can cause Lévy-walk-like movement patterns in organism like humans with advanced learning and reasoning abilities.

One limitation of our simulation is that it was conducted with only two agents. In the future, we aim to run simulations with a large number of decision-making agents to analyse their behaviour in a group and evaluate swarm behaviour.

## V. METHODS

### Extended Bayesian Inference

This section provides an overview of extended Bayesian inference (24, 25), which incorporates the effects of forgetting and learning by introducing causal inference into Bayesian inference. See the SI Appendix for details.

First, we describe the discrete version of Bayesian inference. In Bayesian inference, a number of hypotheses $h_k$ are first defined, and a model for each hypothesis (the generation distribution of data $d$) is prepared in the form of conditional probability $P(d \mid h_k)$. This conditional probability is referred to as the likelihood when the data are fixed and is considered to be a function of the hypothesis. In addition, the confidence $P(h_k)$ for each hypothesis is prepared as a prior probability.

If the confidence at time *t* is $P^t(h_k)$ and we observe data $d^t$, then the posterior probability $P^t(h_k \mid d^t)$ is calculated using Bayes' theorem as follows:

$$P^t(h_k \mid d^t) = \frac{P^t(h_k)P(d^t \mid h_k)}{P^t(d^t)}. \qquad [2]$$

Here, $P^t(d^t)$ is the marginal probability of the data at time *t*, defined as follows:

$$P^t(d^t) = \sum_k P^t(h_k)P(d^t \mid h_k). \qquad [3]$$

The following Bayesian update replaces the posterior probability with the confidence at the next time step.

$$P^{t+1}(h_k) \leftarrow P^t(h_k \mid d^t). \qquad [4]$$

Equations [2] and [4] can be summarized as follows:

$$P^{t+1}(h_k) \leftarrow \frac{P^t(h_k)P(d^t \mid h_k)}{P^t(d^t)}. \quad [5]$$

The estimation proceeds by updating the confidence in each hypothesis according to equation [5] each time the data are observed. Note that in this process, the confidence $P^t(h_k)$ in each hypothesis changes over time, but the model $P(d \mid h_k)$ for each hypothesis does not change.

Given the recursive nature of $P^t(h_k)$, equation [5] can be rewritten as follows:

$$P^{t+1}(h_k) \leftarrow P^1(h_k)\prod_{i=1}^{t}\frac{P(d^i \mid h_k)}{P^i(d^i)}. \quad [6]$$

Here, $P^i(d^i)$ is common to all hypotheses and can be considered a constant. Therefore, if the normalisation process is omitted, equation [6] can be written as follows:

$$P^{t+1}(h_k) \leftarrow P^1(h_k)\prod_{i=1}^{t}P(d^i \mid h_k). \quad [7]$$

The current confidence of each hypothesis is proportional to the prior probability multiplied by the likelihood of the data observed so far at each time step.

Next, we introduce the function of forgetting into Bayesian inference. In order to distinguish between Bayesian inference and extended Bayesian inference, we use C instead of P as follows:

$$C^{t+1}(h_k) \leftarrow \left[\frac{C^t(h_k)}{C^t(d^t)}\right]^{1-\beta} C^t(d^t \mid h_k). \quad [8]$$

Here, β is the forgetting rate (discount rate), and when β=0, equation [8] agrees with equation [5]. If we focus on the recursive nature of $C^t(h_k)$, equation [8] can be transformed as follows:

$$C^{t+1}(h_k) \leftarrow \left[C^1(h_k)\right]^{(1-\beta)^t}\prod_{i=1}^{t}\frac{\left[C^i(d^i \mid h_k)\right]^{(1-\beta)^{t-i}}}{\left[C^i(d^i)\right]^{(1-\beta)^{t+1-i}}}. \quad [9]$$

In equation [9], the denominator $C^i(d^i)$ of the right-hand side is common in each hypothesis and can be considered as a constant, so if the normalisation process is omitted, it can be written as follows:

$$C^{t+1}(h_k) \leftarrow \left[C^1(h_k)\right]^{(1-\beta)^t}\prod_{i=1}^{t}\left[C^i(d^i \mid h_k)\right]^{(1-\beta)^{t-i}}. \quad [10]$$

In other words, the present confidence is multiplied by the past likelihoods with a weaker weight depending on each likelihood's age. When β=0, $C^{t+1}\left(h_k\right)\leftarrow C^1\left(h_k\right)C^1\left(d^1\mid h_k\right)C^2\left(d^2\mid h_k\right)\cdots C^t\left(d^t\mid h_k\right)$, as in Bayesian inference, and the present and distant past likelihoods are evaluated equally.

In contrast, when β = 1, $C^{t+1}\left(h_k\right)\leftarrow C^t\left(d^t\mid h_k\right)$ the next confidence is calculated using only the likelihood for the current observation data.

Next, we introduce a learning effect (inverse Bayesian inference) into Bayesian inference.

$$C^{t+1}\left(d^t\mid h_k\right)\leftarrow\begin{cases}\left[\dfrac{C^t\left(h_k\right)}{C^t\left(d^t\right)}\right]^{\gamma}C^t\left(d^t\mid h_k\right) & \text{if } h_k=\arg\max_{h_i}C^t\left(h_i\right)\\ C^t\left(d^t\mid h_k\right) & \text{otherwise}\end{cases}. \quad [11]$$

Here, we modify the model of the hypothesis with the highest confidence at that point in time (henceforth denoted as $h_{\max}^t$ ) based on the observational data by introducing a learning rate γ. When γ=0, $C^{t+1}\left(d^t\mid h_k\right)\leftarrow C^t\left(d^t\mid h_k\right)$ for any $h_k$ and the model for each hypothesis becomes invariant, as in Bayesian inference. Thus, the extended Bayesian inference adds the forgetting rate β and the learning rate γ to Bayesian inference, and agrees with Bayesian inference when β = γ = 0. The extended Bayesian inference updates the confidence of each hypothesis by equation [10] each time the data are observed and modifies the model of the hypothesis with the maximum confidence by equation [11].

**Imitation game**

As a minimal model of a non-stationary environment, we consider an imitation game with two decision-making agents (Agent1 and Agent2), as shown in Figure 1D. This is a game in which each agent estimates the other's internal state; one can be regarded as the environment of the other. In the imitation game, the target of estimation for each agent is the generative model of the opponent, which changes over time depending on the estimation status of the agent and its opponent. We analyse the behaviour of the agents by running simulations of this game with different forgetting and learning rates.

Let $\{h_1,h_2,\cdots,h_{11}\}$ be the agent's hypotheses. That is, the total number of hypotheses is $K=11$. The initial value of confidence for each hypothesis is set to an equal probability, i.e., $C^1\left(h_k\right)=1/K$ . The generative model for each hypothesis is a normal distribution, and the variance of the normal distribution was fixed at 0.25, and only the mean value was estimated from the observed data. The initial value of the mean of each model was set to $\mu_k^1=\left(k-1\right)/\left(K-1\right)-0.5$ such that these values were equally spaced in the range −0.5 to 0.5.

Each agent samples a real number $d^t$ from the generative model $C^t\left(d \mid h^t_{\max}\right)$ of the hypothesis $h^t_{\max}$ which it believes most at each time step and presents it to its opponent. Each agent observes the real number presented by its counterpart and modifies its generative model and confidence values for hypotheses using the extended Bayesian inference described above.

The simulation was conducted up to 2000 steps, and the data in the interval of $1000 \le t \le 2000$ was used for the analysis.

## AUTHOR CONTRIBUTIONS STATEMENT

S. S. conceived the extended Bayesian inference and the imitation game, conducted the simulation and the analysis, wrote the manuscript. N. M., Y. N., Y. P. G., T. M., H. O., S. M. and U. I. C. contributed to the interpretation of the study findings. All authors participated in the editing and revision of the final version of the manuscript.

## Acknowledgments

This research is supported by the Center of Innovation Program from the Japan Science and Technology Agency, JST and by JSPS KAKENHI Grant Numbers JP16K01408.

## Competing interests

The SoftBank Robotics Group Corp. provided support in the form of salary for author [N. M.], but did not have any additional role in the study design, data collection and analysis, decision to publish, or preparation of the manuscript. The other authors declare no competing interest.

## Data sharing plans (including all data, documentation, and code used in analysis).

All the data and code used in analysis are planned to be made available in Dryad.

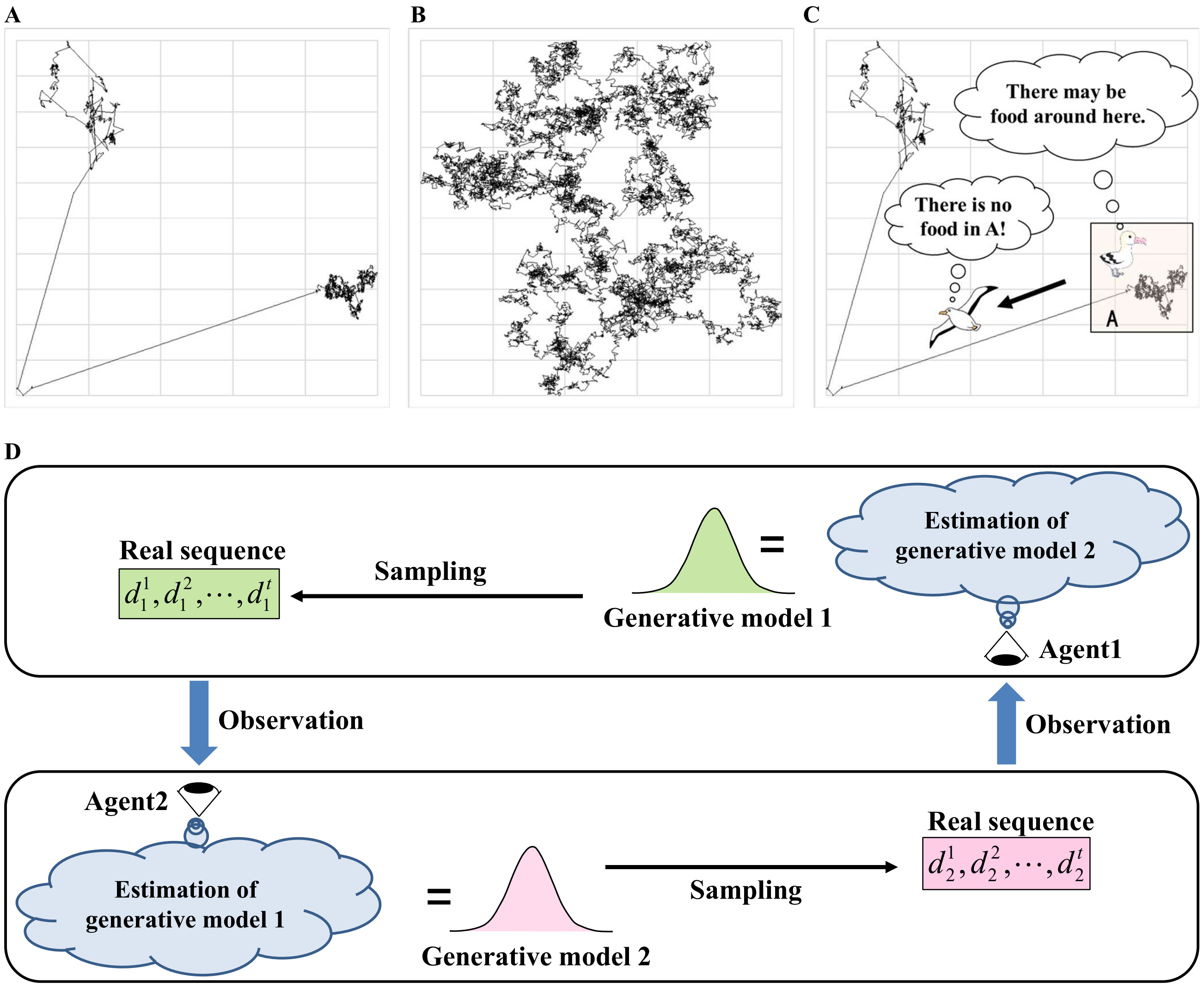

**Fig. 1.** Movement patterns and decision-making. (A) Lévy walk. (B) Brownian walk. (C) Exploratory behaviour based on decision-making. (D) Outline of imitation game with two decision-making agents using Bayesian inference. First, each agent estimates the generative model of the partner agent from the observation data and uses it as its own generative model. Next, each agent samples a real number from its own generative model and presents it to the other party. The agents repeat these procedures in each step.

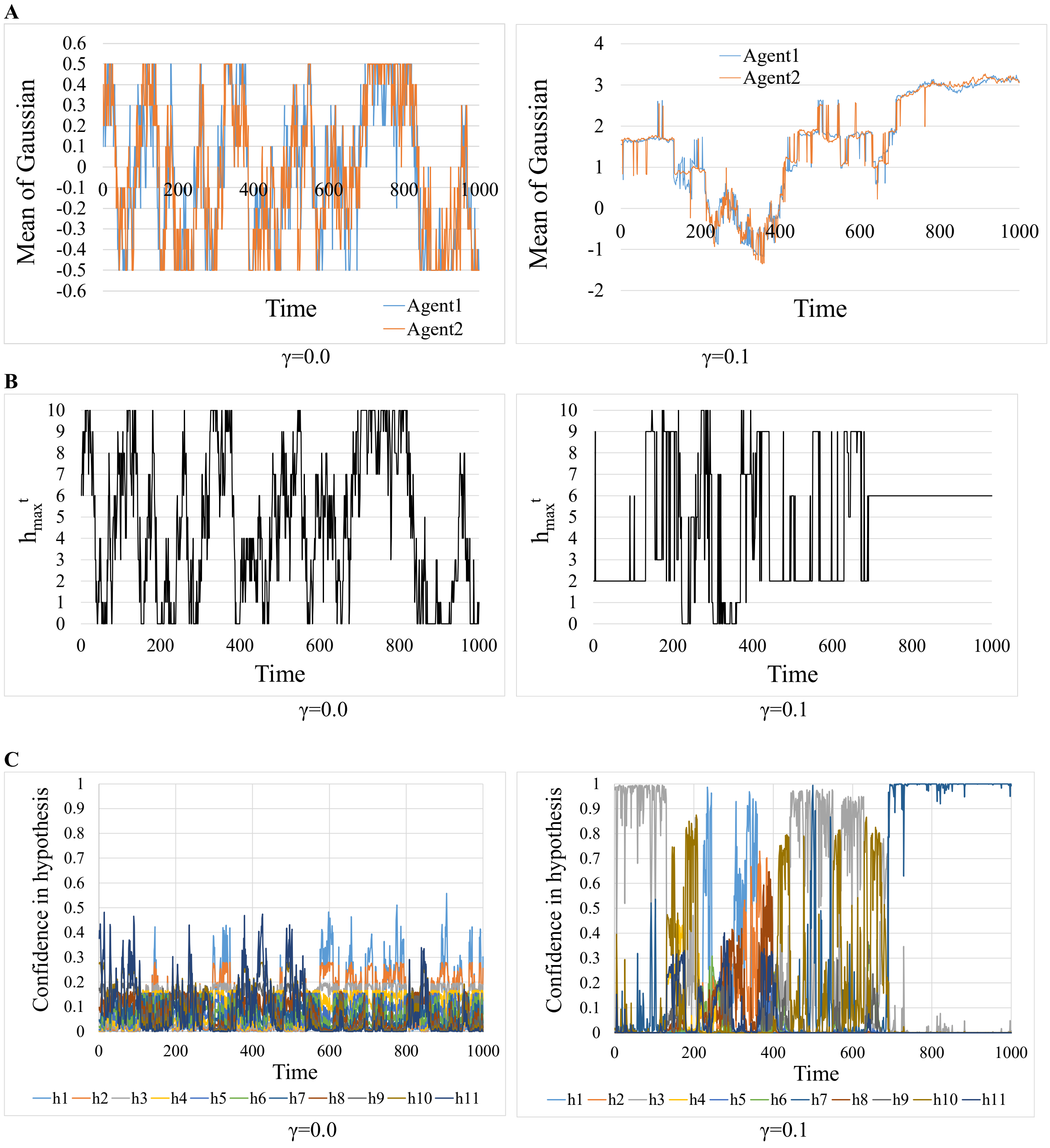

**Fig. 2.** Examples of time evolution when the forgetting rate β=0.3. The left and right columns represent the results in the case without learning (γ=0.0) and the case with learning (γ=0.1), respectively. (A) The normal distribution mean value estimated by each agent. (B) The hypothesis $h_{max}^{t}$ that Agent1 believes the most. (C) Agent1's confidence in each hypothesis.

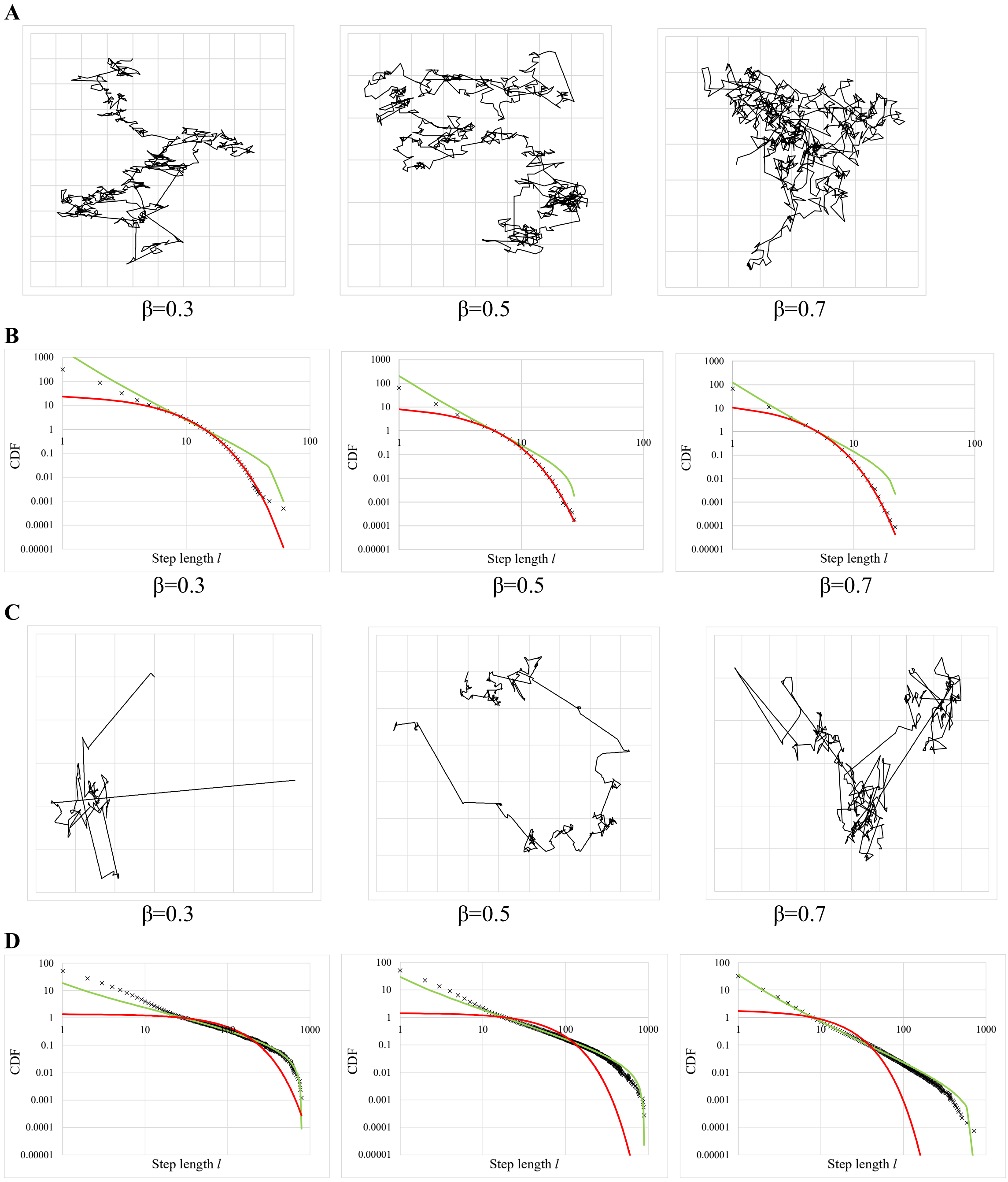

**Fig. 3.** The results of cases in which the forgetting rate was set to β=0.3, 0.5, or 0.7. (A) Examples of two-dimensional movement patterns derived from the time evolution of $h^t_{\max}$ in the case without learning (γ=0.0). These patterns are all Brownian walks. (B) Cumulative distribution function (CDF) of step length $l$ in the case without learning (γ=0.0). The exponents of the exponential distribution for β=0.3, 0.5, and 0.7 were λ=0.24, 0.42, and 0.60, respectively. The minimum $\hat{l}_{\min}$ of fitting data for β=0.3, 0.5, and 0.7 were 14, 6, and 5, respectively. (C) Examples of two-dimensional movement patterns derived from the time evolution of $h^t_{\max}$ in the case with learning (γ=0.1). These patterns are all Lévy walks. (D) Cumulative distribution function (CDF) of step length $l$ in the case with learning (γ=0.1). The exponents of the truncated power law distribution for β=0.3, 0.5, and 0.7 were η=1.73, 2.00, and 2.44, respectively. $1 < \eta \le 3$ is satisfied in all cases. The fitting ranges $\left[\hat{l}_{\min}, \hat{l}_{\max}\right]$ for β=0.3, 0.5, and 0.7 were [29,920], [18,890] and [8,720], respectively.

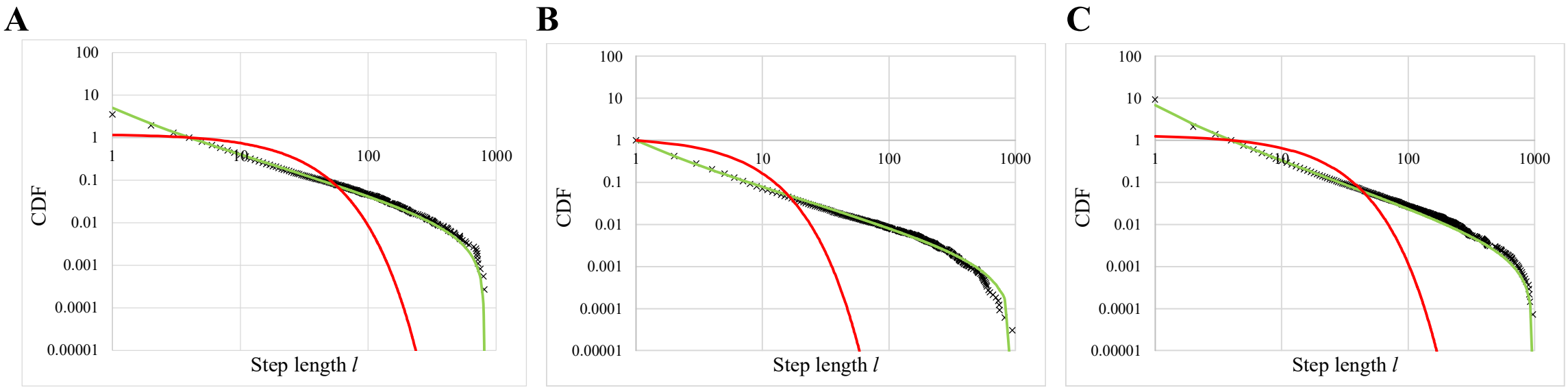

$\alpha$ =0.3 $\alpha$ =0.5 $\alpha$ =0.7

**Fig. 4.** Universally seen Lévy walks. Lévy walks are found in a wide range of parameter regions; the exponents of the truncated power law distribution are close to 2. (A) $\alpha$ =0.3, that is, β=γ=0.3. The exponent of the truncated power law distribution was η=1.91. The fitting range $\left[\hat{l}_{\min}, \hat{l}_{\max}\right]$ was [4,808]. (B) $\alpha$ =0.5, that is, β=γ=0.5. The exponent of the truncated power law distribution was η=1.92. The fitting range $\left[\hat{l}_{\min}, \hat{l}_{\max}\right]$ was [1,942]. (C) $\alpha$ =0.7, that is, β=γ=0.7. The exponent of the truncated power law distribution was η=2.10. The fitting range $\left[\hat{l}_{\min}, \hat{l}_{\max}\right]$ was [4,970].

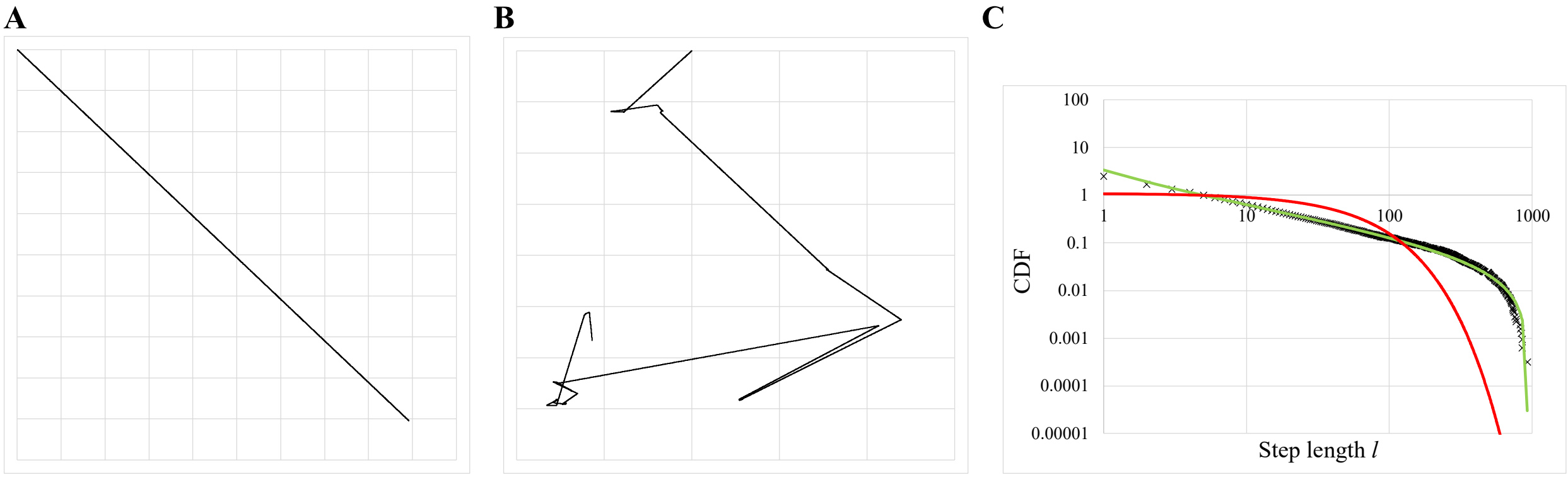

**Fig. 5.** Lévy walks appearing with changes in forgetting rates. Lévy walks are found in the parameter region between the ballistic pattern and the Brownian walk in the case without learning (γ=0.0). (A) The movement pattern when β=0.0, γ=0.0 (Bayesian inference). The pattern is a straight line. (B) The movement pattern when β=0.005, γ=0.0. The pattern is a Lévy walk. (C) Cumulative distribution function (CDF) of step length *l* when β=0.005, γ=0.0. The exponent of the truncated power law distribution was η=1.6. The fitting range $\left[\hat{l}_{\min}, \hat{l}_{\max}\right]$ was [5,927].

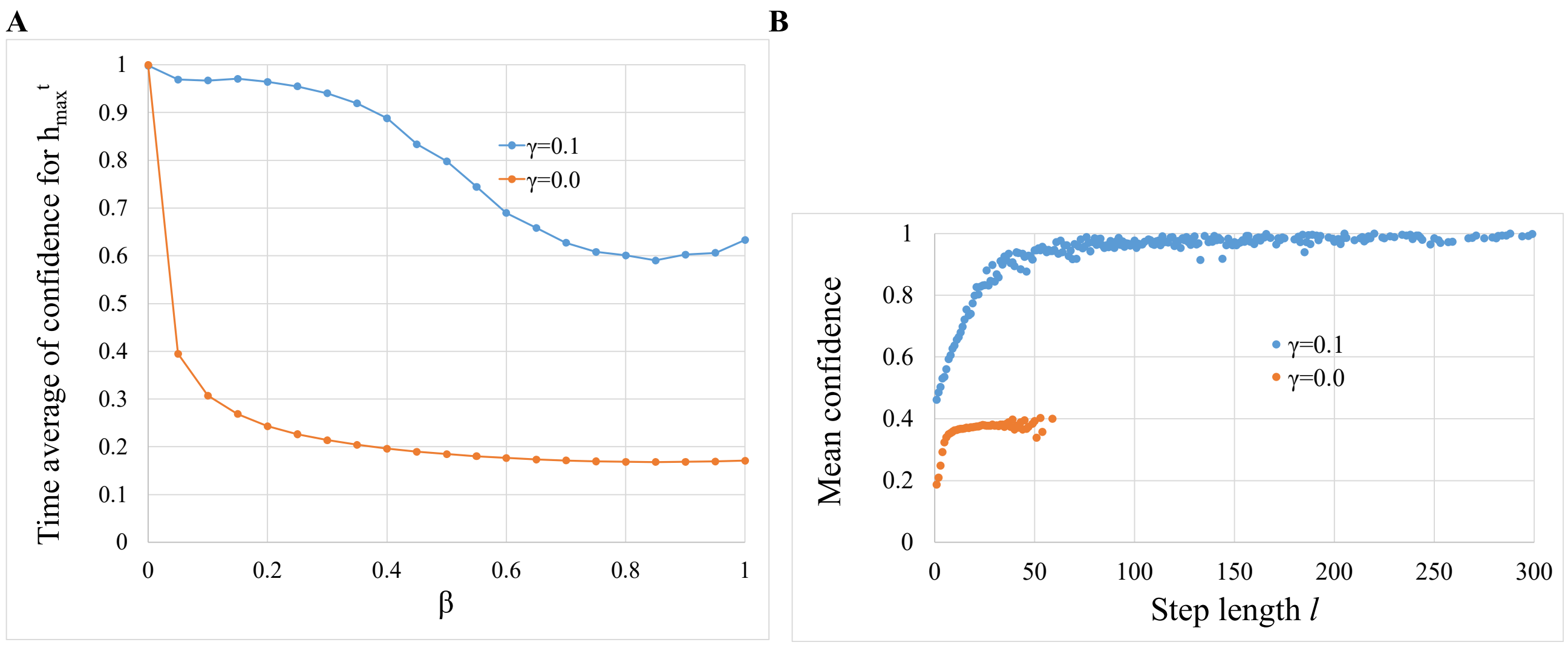

**Fig. 6.** Comparison between cases with no learning (γ = 0.0) and learning (γ = 0.1). (A) Change in the time average of confidence level with change in the forgetting rate. (B) The relationship between step length and confidence (β=0.3). The vertical axis represents the average confidence during the movement of the step length indicated by the horizontal axis. As the step length increases, the confidence in the movement increases. The step length on the horizontal axis is displayed up to 300 for clarity.

# Supplementary Information for

## Lévy walks derived from a Bayesian decision-making model in non-stationary environments

**Shuji Shinohara[1*], Nobuhito Manome[12], Yoshihiro Nakajima[3], Yukio Pegio Gunji[4], Toru Moriyama[5], Hiroshi Okamoto[1], Shunji Mitsuyoshi[1], Ung-il Chung[1]**

1. Department of Bioengineering, Graduate School of Engineering, The University of Tokyo, Tokyo, 113-8656, Japan
2. Department of Research and Development, SoftBank Robotics Group Corp., Tokyo, 105-0021, Japan
3. Graduate School of Economics, Osaka City University, Osaka, 558-8585, Japan
4. Department of Intermedia Art and Science, School of Fundamental Science and Technology, Waseda University, Tokyo, 169-8555, Japan
5. Faculty of Textile Science, Shinshu University, Ueda, 386-8567, Japan

**Extended Bayesian Inference**

In the field of cognitive psychology, many causal induction experiments have been conducted to determine how humans evaluate the strength of the causal relationship between two events (1-5). In the case of the usual conditional statement 'if *p*, then *q*', one would think that the confidence in this statement would be proportional to the probability $P(q \mid p)$ of *q* occurring after *p* occurs (6).

In contrast, it has been experimentally demonstrated that humans have a strong sense of causal relation between a cause *p* and an effect *q* when $P(p \mid q)$ is high as well as when $P(q \mid p)$ is high. Specifically, the causal intensity that people feel between *p* and *q* can be approximated by the geometric mean of $P(q \mid p)$ and $P(p \mid q)$. This is called the dual-factor heuristics (DFH) model (1). If the causal intensity between *p* and *q* is denoted as $DFH(q \mid p)$, then $DFH(q \mid p) = \sqrt{P(q \mid p)P(p \mid q)}$. Here, note that $DFH(q \mid p) = DFH(p \mid q)$ is valid. Such an inference is called 'symmetry inference'.

More generally, Shinohara et al. proposed the following model to express the causal strength $C(q \mid p)$ between *p* and *q* as the generalised weighted average of $P(q \mid p)$ and $P(p \mid q)$ (7).

$$C(q \mid p) = \left[\alpha P(q \mid p)^m + (1-\alpha) P(p \mid q)^m\right]^{1/m} \quad [S1]$$

The generalised weighted average of the variables $x$ and $y$ is expressed by the following equation using the parameters $\alpha$ and $m$.

$$\mu(\alpha, m) = \left[(1-\alpha) x^m + \alpha y^m\right]^{1/m} \quad [S2]$$

Here, $\alpha$ takes a value of the range $0.0 \le \alpha \le 1.0$ and represents the weighting of $x$ and $y$. $m$ takes a value of the range $-\infty \le m \le \infty$ and represents the way the average is taken. For example, if $\alpha = 0.5$ and $m = 1.0$, then $\mu(0.5, 1.0) = 0.5x + 0.5y$, which represents the arithmetic mean. If $\alpha = 0.5$ and $m = -1.0$, then $\mu(0.5, -1.0) = 2xy/(x+y)$, which represents the harmonic mean. Although equation [S2] cannot be defined when $m = 0.0$, if we denote the mean value in the limit of $m \to 0.0$ as $\mu(\alpha, 0.0)$, then we have $\mu(\alpha, 0.0) = x^{1-\alpha} y^{\alpha}$ and the geometric mean $\mu(0.5, 0.0) = \sqrt{xy}$ when $\alpha = 0.5$. If we set $\alpha = 0.0$ here, we obtain $C(q \mid p) = P(q \mid p)$ regardless of the value of *m*, and *C* corresponds to the conditional probability *P*.

Furthermore, Shinohara et al. proposed an extended Bayesian inference that incorporates such a causal inference element into Bayesian inference (7, 8).

$$C^{t+1}(h_k) \leftarrow \left[\alpha C^t(d^t \mid h_k)^m + (1-\alpha) C^t(h_k \mid d^t)^m\right]^{1/m} = \frac{C^t(h_k) C^t(d^t \mid h_k)}{\left[(1-\alpha) C^t(d^t)^{-m} + \alpha C^t(h_k)^{-m}\right]^{-1/m}} \quad [S3]$$

$$C^{t+1}(d^t \mid h_k) \leftarrow \left[(1-\alpha) C^t(d^t \mid h_k)^m + \alpha C^t(h_k \mid d^t)^m\right]^{1/m} = \frac{C^t(h_k) C^t(d^t \mid h_k)}{\left[(1-\alpha) C^t(h_k)^{-m} + \alpha C^t(d^t)^{-m}\right]^{-1/m}} \quad [S4]$$

Here,

$$C^t(d^t) = \sum_k C^t(h_k) C^t(d^t \mid h_k) \quad [S5]$$

In equation [S3], we omit the description of the normalisation process to make the confidence a probability. If we set $\alpha = 0$ in equation [S3], we obtain the same form as the case of Bayesian inference.

$$C^{t+1}(h_k) \leftarrow \frac{C^t(h_k) C^t(d^t \mid h_k)}{C^t(d^t)} \quad [S6]$$

When $\alpha = 0$, equation [S4] is expressed as follows, and the model of the hypothesis is invariant.

$$C^{t+1}(d^t \mid h_k) \leftarrow C^t(d^t \mid h_k) \quad [S7]$$

That is, equation [S4] is greatly reduced and the extended Bayesian inference agrees with the Bayesian inference.

On the other hand, when $\alpha > 0$, the model is deformed by equation [S4]. In this paper, we do not update the models of all the hypotheses, but only the model of the hypothesis $h^t_{\max}$ that has the highest confidence at that time.

$$C^{t+1}\left(d^t \mid h^t_{\max}\right) \leftarrow \frac{C^t\left(h^t_{\max}\right) C^t\left(d^t \mid h^t_{\max}\right)}{\left[(1-\alpha) C^t\left(h^t_{\max}\right)^{-m} + \alpha C^t\left(d^t\right)^{-m}\right]^{-1/m}} \quad [S8]$$

If there are multiple hypotheses with an equally high maximum degree of confidence, one of them is selected at random.

For simplicity, we have fixed $m = 0$ this in this paper. When $m = 0$, equation [S3] can be transformed as follows:

$$C^{t+1}\left(h_k\right) \leftarrow \frac{C^t\left(h_k\right) C^t\left(d^t \mid h_k\right)}{C^t\left(d^t\right)^{1-\alpha} C^t\left(h_k\right)^{\alpha}} = \left[\frac{C^t\left(h_k\right)}{C^t\left(d^t\right)}\right]^{1-\alpha} C^t\left(d^t \mid h_k\right) \quad [S9]$$

Noting the recurrent nature of $C^t\left(h_k\right)$, equation [S9] can be further transformed as follows:

$$C^{t+1}\left(h_k\right) \leftarrow \left[C^1\left(h_k\right)\right]^{(1-\alpha)^t} \prod_{i=1}^{t} \frac{\left[C^i\left(d^i \mid h_k\right)\right]^{(1-\alpha)^{t-i}}}{\left[C^i\left(d^i\right)\right]^{(1-\alpha)^{t+1-i}}} \quad [S10]$$

In equation [S10], the denominator $C^t\left(d^t\right)$ of the right-hand side is common in each hypothesis and can be considered as a constant, so if the normalisation process is omitted, it can be written as follows:

$$C^{t+1}\left(h_k\right) \leftarrow \left[C^1\left(h_k\right)\right]^{(1-\alpha)^t} \prod_{i=1}^{t} \left[C^i\left(d^i \mid h_k\right)\right]^{(1-\alpha)^{t-i}} \quad [S11]$$

When $m = 0$, equation [S8] can be transformed with respect to $h^t_{\max}$ as follows:

$$C^{t+1}\left(d^t \mid h^t_{\max}\right) \leftarrow \frac{C^t\left(h^t_{\max}\right) C^t\left(d^t \mid h^t_{\max}\right)}{C^t\left(h^t_{\max}\right)^{1-\alpha} C^t\left(d^t\right)^{\alpha}} = \left[\frac{C^t\left(h^t_{\max}\right)}{C^t\left(d^t\right)}\right]^{\alpha} C^t\left(d^t \mid h^t_{\max}\right) \quad [S12]$$

Through the processes described above, the confidence values for each hypothesis and the model for the hypothesis with maximum confidence are corrected whenever the data are observed. We refer to the latter process of modifying the model for $h^t_{\max}$ as inverse Bayesian inference (9). If the former process of updating the confidence values for hypotheses is referred to as inference, inverse Bayesian inference can be called 'learning' because it forms a model for a hypothetical instead of an inference. In this sense, although $\alpha$ in equation [S9] and $\alpha$ in equation [S12] are the same parameter, they can be considered as forgetting and learning rates, respectively, and can be kept separate. In this paper, we introduce a forgetting rate β and a learning rate γ, and transform equations [S9] and [S12] as follows:

$$C^{t+1}(h_k) \leftarrow \frac{C^t(h_k) C^t(d^t \mid h_k)}{C^t(d^t)^{1-\beta} C^t(h_k)^{\beta}} = \left[\frac{C^t(h_k)}{C^t(d^t)}\right]^{1-\beta} C^t(d^t \mid h_k) \qquad \text{[S13]}$$

$$C^{t+1}(d^t \mid h^t_{\max}) \leftarrow \frac{C^t(h^t_{\max}) C^t(d^t \mid h^t_{\max})}{C^t(h^t_{\max})^{1-\gamma} C^t(d^t)^{\gamma}} = \left[\frac{C^t(h^t_{\max})}{C^t(d^t)}\right]^{\gamma} C^t(d^t \mid h^t_{\max}) \qquad \text{[S14]}$$

See the next section for methods that apply the normal distribution as a specific generative model in the extended Bayesian inference.

**Applying a normal distribution**

In our model, the confidences $C^t(h_k)$ for each hypothesis $h_k$ and the model $C^t(d^t \mid h^t_{\max})$ for the hypothesis $h^t_{\max}$ with maximum confidence are corrected whenever the data $d^t$ are observed at time *t* using equations [S13] and [S14].

In this paper, we consider the following one-dimensional normal distribution as a concrete model of the hypothesis. For simplicity, we assume that the variance $\Sigma$ is the same at all times for all hypotheses, and we consider only the difference in the mean $\mu^t_k$. See (8) for a method to estimate the mean and variance simultaneously.

$$C^t(d \mid h_k) = N(d \mid \mu^t_k, \Sigma) = \frac{1}{\sqrt{2\pi\Sigma}} \exp\left[-\frac{(d-\mu^t_k)^2}{2\Sigma}\right] \qquad \text{[S15]}$$

When adopting a normal distribution as a model, if the number of hypotheses is discrete and finite, $C^t(h_k)$ is a probability and $C^t(d \mid h_k)$ or $C^t(d)$ is a probability density. For this reason, we introduce a positive number $\Delta$ when computing equations [S13] and [S14] as follows:

$$C^{t+1}(h_k) \leftarrow \left[C^t(h_k)\right]^{1-\beta} \Delta N(d^t \mid \mu^t_k, \Sigma) \qquad \text{[S16]}$$

$$\begin{aligned} C^{t+1}(d^t \mid h^t_{\max}) &\leftarrow \frac{1}{\Delta}\left[\frac{C^t(h^t_{\max})}{\sum_k C^t(h_k) \Delta N(d^t \mid \mu^t_k, \Sigma)}\right]^{\gamma} \Delta N(d^t \mid \mu^t_{\max}, \Sigma) \\ &= \frac{1}{\Delta^{\gamma}}\left[\frac{C^t(h^t_{\max})}{\sum_k C^t(h_k) N(d^t \mid \mu^t_k, \Sigma)}\right]^{\gamma} N(d^t \mid \mu^t_{\max}, \Sigma) \end{aligned} \qquad \text{[S17]}$$

Here, $\mu^t_{\max}$ is the mean of the model of hypothesis $h^t_{\max}$.

In equation [S16], the term $\Delta$ is common to all hypotheses and can be cancelled by normalisation. Therefore, if we omit the normalisation process, we can express [S16] as follows:

$$C^{t+1}(h_k) \leftarrow \left[C^t(h_k)\right]^{1-\beta} N\left(d^t \mid \mu_k^t, \Sigma\right) \qquad \text{[S18]}$$

In equation [S18], once the confidence $C^t(h_k)$ of each hypothesis becomes 0, it is always 0 thereafter. To prevent this, a normalisation process (smoothing) is performed by adding a small positive constant $\varepsilon$ to the confidence of each hypothesis obtained by Equation [S18].

$$C^{t+1}(h_k) \leftarrow \frac{C^{t+1}(h_k)+\varepsilon}{\sum_{j=1}^{K}\left[C^{t+1}(h_j)+\varepsilon\right]} = \frac{C^{t+1}(h_k)+\varepsilon}{K\varepsilon + \sum_{j=1}^{K} C^{t+1}(h_j)} \text{[S19]}$$

In this paper, we set $\varepsilon = 10^{-8}$. $K$ represents the number of hypotheses.

When observing the data $d^t$ at time $t$, the likelihood is changed to $C^{t+1}\left(d^t \mid h_{\max}^t\right)$ by equation [S17]. Accordingly, we modify the mean of the model of the hypothesis $h_{\max}^t$ from $\mu_{\max}^t$ to $\mu_{\max}^{t+1}$ so that the following equation is satisfied:

$$C^{t+1}\left(d^t \mid h_{\max}^t\right) = N\left(d^t \mid \mu_{\max}^{t+1}, \Sigma\right) = \frac{1}{\sqrt{2\pi\Sigma}} \exp\left[-\frac{\left(d^t - \mu_{\max}^{t+1}\right)^2}{2\Sigma}\right] \qquad \text{[S20]}$$

Solving equation [S20] for $\mu_{\max}^{t+1}$ yields the following two solutions.

$$\begin{aligned} \mu_1 &= d^t + \sqrt{-2\Sigma \log\left[C^{t+1}\left(d^t \mid h_{\max}^t\right)\sqrt{2\pi\Sigma}\right]} \\ \mu_2 &= d^t - \sqrt{-2\Sigma \log\left[C^{t+1}\left(d^t \mid h_{\max}^t\right)\sqrt{2\pi\Sigma}\right]} \end{aligned} \qquad \text{[S21]}$$

We define $\mu_{\max}^{t+1}$ as the solution that is closer to $\mu_{\max}^t$. Specifically,

$$\mu_{\max}^{t+1} = \begin{cases} \mu_1 & if \quad \left|\mu_1 - \mu_{\max}^t\right| \le \left|\mu_2 - \mu_{\max}^t\right| \\ \mu_2 & otherwise \end{cases} \qquad \text{[S22]}$$

However, in order to solve equation [S20] as an equation of $\mu_{\max}^{t+1}$, $C^{t+1}\left(d^t \mid h_{\max}^t\right)$ must be in the range of $0 < C^{t+1}\left(d^t \mid h_{\max}^t\right) \le \frac{1}{\sqrt{2\pi\Sigma}}$.

For this reason, we set the following constraint when calculating $C^{t+1}\left(d^t \mid h_{\max}^t\right)$ using equation [S17]:

$$C^{t+1}\left(d^t \mid h_{\max}^t\right) \leftarrow \min\left(\max\left(C^{t+1}\left(d^t \mid h_{\max}^t\right), \varepsilon\right), \frac{1}{\sqrt{2\pi\Sigma}}\right) \text{[S23]}$$

We set $\varepsilon = 10^{-8}$.

Let us consider a situation $C^t\left(h_{\max}^t\right) \approx 1$ where the confidence of the hypothesis with the highest confidence is almost 1. Because the confidence of any other hypothesis other than $h_{\max}^t$ is almost zero by the constraint of $\sum_k C^t(h_k) = 1$, we obtain $C^t\left(d^t\right)=\sum_k C^t\left(h_k\right)C^t\left(d^t \mid h_k\right)\approx C^t\left(d^t \mid h^t_{\max}\right)$ from $C^t\left(d^t\right)=\sum_k C^t\left(h_k\right)C^t\left(d^t \mid h_k\right)$. Therefore, equation [S17] can be transformed as follows:

$$C^{t+1}\left(d^t \mid h^t_{\max}\right)\leftarrow \frac{1}{\Delta^\gamma}\left[\frac{C^t\left(h^t_{\max}\right)}{\sum_k C^t\left(h_k\right)N\left(d^t \mid \mu^t_k,\Sigma\right)}\right]^\gamma N\left(d^t \mid \mu^t_{\max},\Sigma\right)\approx\left(\frac{1}{\Delta}\right)^\gamma\left[C^t\left(d^t \mid h^t_{\max}\right)\right]^{1-\gamma} \quad [S24]$$

If equation [S24] is denoted by $x^{t+1}=f\left(x^t\right)=\left(\frac{1}{\Delta}\right)^\gamma\left(x^t\right)^{1-\gamma}$, then $f\left(x^t\right)$ becomes a concave function.

Solving $x^t=f\left(x^t\right)$ results in $x^t=0,\frac{1}{\Delta}$.

The fixed point $\left(x^t,f\left(x^t\right)\right)=\left(\frac{1}{\Delta},\frac{1}{\Delta}\right)$ is a stable point because $x^t\geq f\left(x^t\right)$ when $x^t>\frac{1}{\Delta}$ and $x^t\leq f\left(x^t\right)$ when $x^t<\frac{1}{\Delta}$. In this study, we set $\Delta=\sqrt{2\pi\Sigma}$. In this case, $C^{t+1}\left(d^t \mid h^t_{\max}\right)$ approaches the vertex $\frac{1}{\Delta}=\frac{1}{\sqrt{2\pi\Sigma}}$ of the normal distribution whenever data $d^t$ are observed.

As shown in formula [S20], $\mu^{t+1}_{\max}$ is determined to satisfy the condition $C^{t+1}\left(d^t \mid h^t_{\max}\right)=N\left(d^t \mid \mu^{t+1}_{\max},\Sigma\right)$. This means that $\mu^{t+1}_{\max}$ approaches the observation data $d^t$.

To summarize the above ideas:

1. Set values for parameters $\beta,\gamma,\varepsilon,K$.
2. Establish initial values for $\Sigma,\mu^1_k,C^1\left(h_k\right)\quad(k=1,2,\cdots K)$.
3. Repeat the following whenever data $d^t$ are observed.
    - Find the hypothesis $h^t_{\max}$ with the maximum confidence.
    - Update the confidence $C^{t+1}\left(h_k\right)$ of each hypothesis using formulas [S18] and [S19].
    - Update the likelihood $C^{t+1}\left(d^t \mid h^t_{\max}\right)$ of the hypothesis $h^t_{\max}$ for the observed data $d^t$ using formula [S17].
    - Correct the mean $\mu^{t+1}_{\max}$ of the model for the hypothesis $h^t_{\max}$ using formulas [S21] and [S22] to match the new likelihood $C^{t+1}\left(d^t \mid h^t_{\max}\right)$.

**Fitting to simulation data**

**Fitting to truncated power law distribution (TP)**

Here, we describe a method to fit the frequency distribution of step length $l$ observed by simulation to the truncated power law distribution (TP). The method was based on references (10-14). Specifically, we want to find the minimum value $\hat{l}_{\min}$ and maximum value $\hat{l}_{\max}$ of the data to be fitted to the TP and the exponent $\hat{\eta}$ of the TP model that best fits the data in the range of $\hat{l}_{\min} \le l \le \hat{l}_{\max}$. First, $\hat{l}_{\max}$ is the longest step length in the observation data. Next, we describe the method of calculating $\hat{l}_{\min}$. In the case of a discrete distribution, the TP in the range of $l_{\min} \le l \le l_{\max}$ is expressed by the following formula:

$$p\left(l;\eta,l_{\min},l_{\max}\right)=\frac{l^{-\eta}}{\zeta\left(\eta,l_{\min},l_{\max}\right)},\quad \zeta\left(\eta,l_{\min},l_{\max}\right)=\sum_{i=l_{\min}}^{l_{\max}} i^{-\eta} \qquad \text{[S25]}$$

The CDF of $p\left(l;\eta,l_{\min},l_{\max}\right)$ is expressed in the following equation.

$$P\left(l;\eta,l_{\min},l_{\max}\right)=\frac{\zeta\left(\mu,l,l_{\max}\right)}{\zeta\left(\mu,l_{\min},l_{\max}\right)} \qquad \text{[S26]}$$

If the observed data in the range of $l_{\min} \le l \le l_{\max}$ are $\left\{l_1,l_2,\cdots,l_n\right\}$, then the log-likelihood of these data for TP is calculated using equation [S25] as follows:

$$L\left(\eta;l_{\min},l_{\max}\right)=\sum_{i=1}^{n}\ln p\left(l_i;l_{\min},l_{\max},\eta\right)=-n\ln\zeta\left(\eta,l_{\min},l_{\max}\right)-\eta\sum_{i=1}^{n}\ln l_i \qquad \text{[S27]}$$

The exponent $\hat{\eta}\left(l_{\min},l_{\max}\right)$ of the TP model that best fits the data in the range of $l_{\min} \le l \le l_{\max}$ is $\eta$, which maximizes $L\left(\eta;l_{\min},l_{\max}\right)$. Specifically, we varied $\eta$ from 0.5 to 3.5 in increments of 0.01 to obtain $\hat{\eta}\left(l_{\min},l_{\max}\right)$, which numerically maximizes equation [S27].

We introduce the Kolmogorov-Smirnov static $D\left(l_{\min},l_{\max}\right)$ to measure the closeness of the cumulative frequency distribution $S\left(l;l_{\min},l_{\max}\right)$ created from the data in the range of $l_{\min} \le l \le l_{\max}$ and the theoretical cumulative frequency distribution $P\left(l;\hat{\eta}\left(l_{\min},l_{\max}\right),l_{\min},l_{\max}\right)$ represented by equation [S26].

$$D\left(l_{\min},l_{\max}\right)=\max_{l_{\min}\le l\le l_{\max}}\left|S\left(l;l_{\min},l_{\max}\right)-P\left(l;\hat{\eta}\left(l_{\min},l_{\max}\right),l_{\min},l_{\max}\right)\right| \qquad \text{[S28]}$$

If we fix $l_{\max}=\hat{l}_{\max}$, then $D\left(l_{\min},\hat{l}_{\max}\right)$ is a function of $l_{\min}$. We numerically choose $l_{\min}$ out of the observed data, which minimizes $D\left(l_{\min},\hat{l}_{\max}\right)$. That is, $\hat{l}_{\min}=\arg\min_{l_{\min}} D\left(l_{\min},\hat{l}_{\max}\right)$. In the above, $\hat{l}_{\min}$ and $\hat{l}_{\min}$ were obtained. Finally, we find the exponent $\hat{\mu}=\hat{\mu}\left(\hat{l}_{\min},\hat{l}_{\max}\right)$ of the TP model that best fits the data in the range of $\hat{l}_{\min} \le l \le \hat{l}_{\max}$ using the formula [S27].

**Fitting to exponential distribution (EP)**

In this section, our goal is to find the minimum value $\hat{l}_{\min}$ of the observed data to be fitted to the exponential distribution (EP) model and the exponent $\hat{\lambda}$ of the EP model that best fits the data in the range of $\hat{l}_{\min} \le l$. In the discrete case, the EP in the range of $l_{\min} \le l$ is expressed in the following equation:

$$p(l;l_{\min},\lambda) = \left(1-e^{-\lambda}\right)e^{-\lambda(l-l_{\min})} \qquad \text{[S29]}$$

The CDF of $p(l;l_{\min},\lambda)$ is expressed as follows:

$$P(l;l_{\min},\lambda) = e^{-\lambda(l-l_{\min})} \qquad \text{[S30]}$$

If the data in the range of $l_{\min} \le l$ is $\{l_1, l_2, \cdots, l_m\}$, then the log likelihood for these data is expressed as:

$$L(\lambda;l_{\min}) = \sum_{i=1}^{m} \ln p(l_i;l_{\min},\lambda) = m\ln\left(1-e^{-\lambda}\right) - \lambda\sum_{i=1}^{m}\left(l_i - l_{\min}\right) \qquad \text{[S31]}$$

The exponent $\hat{\lambda}(l_{\min})$ that maximizes $L(\lambda;l_{\min})$ is found as a solution to $\frac{\partial L(\lambda;l_{\min})}{\partial \lambda} = 0$ by the following formula:

$$\hat{\lambda}(l_{\min}) = \ln\left(\frac{m}{\sum_{i=1}^{m}\left(l_i - l_{\min}\right)} + 1\right) \qquad \text{[S32]}$$

$\hat{l}_{\min}$ is calculated from the simulation data and $D_{adj}$ obtained from equation [S30], as in the case of TP. The final value is $\hat{\lambda} = \hat{\lambda}\left(\hat{l}_{\min}\right)$.

**Comparison of truncated power law distribution (TP) and exponential distribution (EP)**

In this section, we describe a method to determine which of the two distribution models, TP or EP, is more suitable for the simulation data. We use Akaike Information Criteria weights (AICw) for comparison (14). First, the Akaike Information Criterion (AIC) for data in the range of $l_{\min} \le l \le l_{\max}$ is defined as follows:

$$\begin{aligned} AIC_{TP} &= -2\ln\left(L\left(\hat{\eta};l_{\min},l_{\max}\right)\right) + 2 \\ AIC_{EP} &= -2\ln\left(L\left(\hat{\lambda};l_{\min}\right)\right) + 2 \end{aligned} \qquad \text{[S33]}$$

The AIC difference $\Delta$ is then calculated as follows:

$$\begin{aligned} AIC_{\min} &= \min\left(AIC_{TP}, AIC_{EP}\right) \\ \Delta_{TP} &= AIC_{TP} - AIC_{\min} \\ \Delta_{EP} &= AIC_{EP} - AIC_{\min} \end{aligned} \quad [S34]$$

Finally, AICw are calculated as follows:

$$\begin{aligned} w_{TP} &= \frac{e^{-\Delta_{TP}/2}}{e^{-\Delta_{TP}/2} + e^{-\Delta_{EP}/2}} \\ w_{EP} &= \frac{e^{-\Delta_{EP}/2}}{e^{-\Delta_{TP}/2} + e^{-\Delta_{EP}/2}} \end{aligned} \quad [S35]$$

First, using the data in the range of $\hat{l}_{\min} \le l \le \hat{l}_{\max}$ calculated during the fitting of the TP, we find the most appropriate exponents $\hat{\eta}$ and $\hat{\lambda}$ for each model.

Next, these exponents are used to calculate and compare AICw. Then, we change $\hat{l}_{\min}$ to the one calculated during the fitting of the EP and compare. If $w_{TP} > w_{EP}$ for both data, the TP is considered to fit the simulated data better. On the other hand, if $w_{TP} < w_{EP}$ for both data, the EP is considered to fit the simulated data better. In case of discrepancies between the results in both data, the following indicators were defined and judged according to reference (10).

$$\begin{aligned} D_{adj,TP} &= \frac{\ln N}{\ln n_{TP}} D_{TP} \\ D_{adj,EP} &= \frac{\ln N}{\ln n_{EP}} D_{EP} \end{aligned} \quad [S36]$$

where $D_{\mathrm{TP}}$ and $D_{\mathrm{EP}}$ are the Kolmogorov-Smirnov static calculated during the model fitting of the TP and EP, respectively. $N$ is the total number of observed data points, and $n_{TP}$ and $n_{EP}$ are the number of observed data points used in each model fitting. In other words, the index considers a model that can fit more observational data to be a better model. In the case of $D_{adj,TP} < D_{adj,EP}$, the TP is considered to fit the simulation data better. Conversely, when $D_{adj,TP} > D_{adj,EP}$, EP is considered to be a better fit to the simulation data. If the optimal model was judged to be a TP, it is considered to be a Lévy walk if $1 < \hat{\eta} \le 3$ was satisfied.